\documentclass{article}

\usepackage{arxiv}

\usepackage[utf8]{inputenc} % allow utf-8 input
\usepackage[T1]{fontenc}    % use 8-bit T1 fonts
\usepackage{hyperref}       % hyperlinks
\usepackage{url}            % simple URL typesetting
\usepackage{booktabs}       % professional-quality tables
\usepackage{amsfonts}       % blackboard math symbols
\usepackage{nicefrac}       % compact symbols for 1/2, etc.
\usepackage{microtype}      % microtypography
\usepackage{lipsum}		% Can be removed after putting your text content
\usepackage{graphicx}
\usepackage{natbib}
\usepackage{amsmath,amssymb,amsfonts,bm}
\usepackage{doi}

\title{Intrinsically motivated learning of  \\ causal world models}

\author{Louis Annabi \\
    U2IS, ENSTA Paris, Institut Polytechnique de Paris, France \\
	\texttt{louis.annabi@gmail.com} \\
}

\renewcommand{\shorttitle}{\textit{arXiv} Template}

\begin{document}
\maketitle

\begin{abstract}
    Despite the recent progress in deep learning and reinforcement learning, transfer and generalization of skills learned on specific tasks is very limited compared to human (or animal) intelligence. The lifelong, incremental building of common sense knowledge might be a necessary component on the way to achieve more general intelligence. A promising direction is to build world models capturing the true physical mechanisms hidden behind the sensorimotor interaction with the environment.
    Here we explore the idea that inferring the causal structure of the environment could benefit from well-chosen actions as means to collect relevant interventional data.
\end{abstract}

% keywords can be removed

\section{Introduction}

Model Based Reinforcement Learning \citep{moerland2020model} techniques allow artificial agents to optimize their behaviors through imagined episodes generated by a learned world model \citep{ha2018world}. Using these methods, the quality of the learned behaviors highly depends on the quality of the agent's world model. In this work, we take inspiration from recent work in causal structure learning to explore two directions of improvement for world model learning: sample efficiency and transfer capabilities. 

It has been shown in \citep{bengio2019meta} that based on the assumption that transfer distributions are obtained with sparse changes with regard to the initial data distribution, it is possible to use the speed of adaptation as an objective to optimize the structure of the internal model of the data. The obtained model should favor a structure composed of different mechanism mimicking the true causal structure of the data. Transferring this idea to RL, world models mimicking the causal structure of the environment (more precisely, the MDP transition function) should generalize better to small changes in the environment. This approach can be related to existing works in the causality literature advocating for the use of interventional data in order to uncover the ground truth structural causal models \citep{scholkopf2021toward}. To this end, the RL setting is interestingly convenient as agent actions can be seen as interventions. By acting, the agent performs interventions that partly affect the state of the environment, which can allow it to uncover some of the causal relations between environment variables.

Regarding the second axis of improvement, sample efficiency, we suggest using a form of intrinsic motivation to guide the causal world model learning. This causal artificial curiosity mechanism should drive the agent towards states (or observations in a partially observable setting) informative about the causal structure of the environment.

The proposed approach rests on two design priors: continuity of the state of the environment, and law of parsimony. Continuity of the state implies that at each time step, the variables composing the environment state evolve smoothly. This has an implication on the causal structure we try to infer, as each state variable must influence its subsequent self. The law of parsimony states that if a variable can be explained with fewer causes, then this explanation should be preferable compared to one based on more causes. In our implementation, this translates into a preference for removing potential causal relations between variables.

\section{Related work}

We take inspiration from some recent works formulating inference of the causal structure as a continuous optimization problem \citep{ke2019learning, brouillard2020differentiable}, and apply those ideas in order to learn causal world models. The question of actively selecting helpful interventions is also being explored in \citep{scherrer2021learning}, where the authors show that targeting the interventions can accelerate the search of the underlying causal structure.

The approach presented in \citep{sontakke2021causal} and labeled causal curiosity is very different from the work presented here. The authors suggest learning courses of actions that facilitates the evaluation of previously identified disentangled variables. In comparison, our work tries to implement intrinsic motivation measures encouraging the exploration of states that are informative about the causal structure of the agent's world model.

\section{Methods}
\label{sec:methods}

This section describes the proposed algorithm for learning the structure and parameters of the neural world model, as well as the intrinsic motivation driving action selection.

\subsection{Model}
\label{sec:model}

\begin{figure}[ht!]
    \centering
    \includegraphics[width=\textwidth]{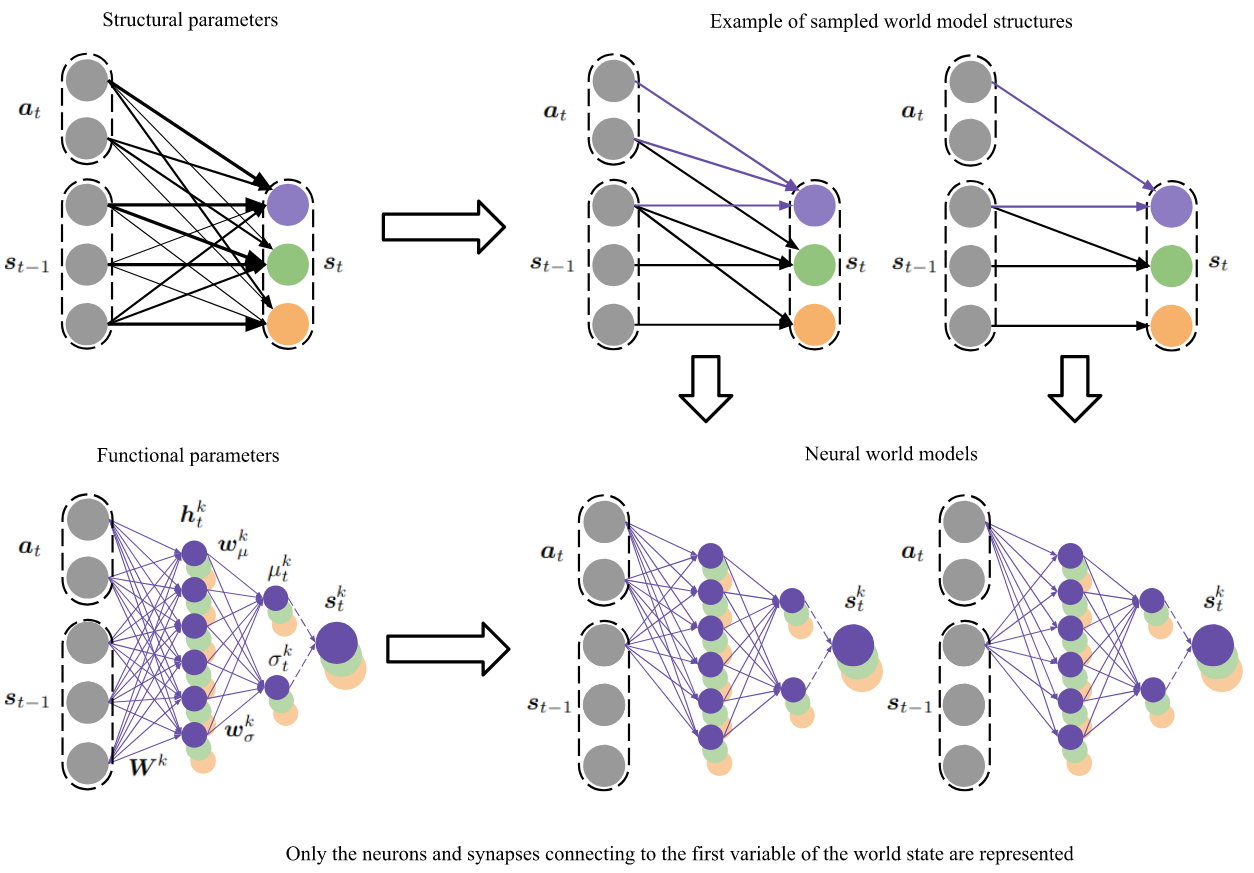}
    \caption{Illustration of the roles of the structural and functional parameters. Top-left: The structural parameters define probabilities of potential cause-effect relations (wider arrows represent more probable relations). Top-right: Two possible world model structures sampled with these structural parameters. Bottom-left: Functional parameters for the prediction of the $k$-th variable of the state $\bm{s}_t$. Bottom-right: Neural networks resulting from the masking of the functional parameters using the sampled world model structures.}
    \label{fig:model}
\end{figure}

We assume that a disentangled representation of the environment state is provided at each time step to the agent, that we denote $\bm{s}_t$. This representation is composed of $d_s$ features that the world model tries to predict based on a past representation $\bm{s}_{t-1}$ (of dimension $d_s$) and current action $\bm{a}_t$ (of dimension $d_a$). We implement this world model as a neural network outputting the mean and standard deviation of a multivariate Gaussian probability distribution onto $\bm{s}_t$:

\begin{equation}
    p(\bm{s}_t ; \theta) = \mathcal{N} \big(\bm{s}_t; \bm{\mu}_\theta(\bm{s}_{t-1}, \bm{a}_t), \bm{\sigma}_\theta(\bm{s}_{t-1}, \bm{a}_t) \big)
\end{equation}

Each feature $k$ of the state $\bm{s}_t$ is estimated based on a set of causes $\mathcal{C}_k$ corresponding to the parent nodes of the feature node $k$ in the causal graph $G$, as represented in figure \ref{fig:model}. As such, this causal graph encodes which features of the past state and current action are responsible for the current state feature $k$. An ideal causal world model should not only provide optimal predictions over the state $\bm{s}_t$, but should also mimic the true causal structure of the environment.

We define two sets of parameters that the agent optimizes during learning: the structural parameters $\bm{\gamma}$ and the functional parameters $\bm{\theta}$. The structural parameters $\bm{\gamma}$ encode the agent's current belief about the causal graph. $sigmoid(\gamma_{ik})$ corresponds to the probability of the input feature $i$ (where $i\leq d_s$ correspond to past state features and $d_s < i \leq d_s+d_a$ correspond to action features) causally affecting the feature $k$ of $\bm{s}_t$. Different possible causal graphs $G$ can be sampled using these parameters. For each possible edge $(i,k)$, we sample the adjacency matrix coefficient $\bm{A}_{ik}$ using the Bernoulli distribution of parameter $sigmoid(\gamma_{ik})$.

The functional parameters are the weights of the neural network implementing the world model. Each feature $k$ of the state is associated with a hidden layer $\bm{h}_t^k$ of dimension $d_h$ as represented in figure \ref{fig:model}. The mean and standard deviation of the prediction over the $k$-th feature are computed as (biases omitted for clarity):

\begin{align}
    \mu^k_t &= \bm{w}_\mu^k \cdot \bm{h}_t^k\\
    \sigma^k_t &= \exp( \bm{w}_\sigma^k \cdot \bm{h}_t^k )
\end{align}

The hidden layers activations are computed based on the input features (action and past state), using the sampled adjacency matrix as mask to ensure that the prediction of feature $k$ is only based on information coming from the parents in the graph $G$.

\begin{equation}
    \bm{h}_t^k = \sigma\big(\bm{W}^k \cdot (\bm{A}_{:k} \odot [ \bm{s}_{t-1} , \bm{a}_t ]) \big)
\end{equation}

where $\sigma$ denotes the sigmoid function, $\bm{A}_{:k}$ denotes the $k$-th column of $\bm{A}$, $\odot$ denotes the element-wise product, and $[\bm{s}_{t-1} , \bm{a}_t]$ denotes the concatenation of the previous state and current action. The functional parameters $\bm{\theta} = (\bm{w}_\mu^k, \bm{w}_\sigma^k, \bm{W}^k)_k\leq d_s$ are shared between the different world model causal structures sampled according to the structural parameters.

\subsection{Learning}
\label{sec:learning}

The functional parameters are optimized in order to increase the log-likelihood of the world model predictions:

\begin{equation}
    \mathcal{L}_G = \log p(\bm{s}; G, \bm{\theta})
\end{equation}

These parameters are found by stochastic gradient descent using the backpropagation algorithm, with dropout caused by the graph sampling. During training, we directly take the means $\bm{\mu}_t$ as states for future prediction and do note use the renormalization trick.

\begin{equation}
    \Delta \bm{\theta} \propto \sum_{G \sim p_{\bm{\gamma}}(G)} \nabla_{\bm{\theta}} \mathcal{L}_G
\end{equation}

The computation graph connecting structural parameters to the log-likelihood is not fully differentiable because of the sampling operation. To optimize the structural parameters, we use the REINFORCE trick:

\begin{equation}
    \Delta \bm{\gamma} \propto \sum_{G \sim p_{\bm{\gamma}}(G)} \nabla_{\bm{\gamma}} \log p_{\bm{\gamma}}(G) \mathcal{L}_G
\end{equation}

In consequence, the structural parameters $\bm{\gamma}$ should be pulled with more strength towards adjacency matrices associated with high prediction accuracy. An issue with this learning algorithm is that it does not penalize structures with useless connections. To avoid this, we add another term to the gradient that favors graphs with fewer edges, acting as a regularization mechanism aligned with Occam's law of parsimony.

\begin{equation}
    \Delta \bm{\gamma} \propto \sum_{G \sim p_{\bm{\gamma}}(G)} \nabla_{\bm{\gamma}} \log p_{\bm{\gamma}}(G) \mathcal{L}_G - \alpha \nabla_{\bm{\gamma}}\sigma(\bm{\gamma})
\end{equation}

where $\alpha$ is an hyperparameter balancing the regularization.

\subsection{Intrinsic motivation}
\label{sec:im}

Having defined the model and the associated learning mechanisms, we now turn to the proposed causal curiosity mechanism. This curiosity should reflect the agent's desire to gather information about the causal structure of the environment. As such, it could for instance reward learning progress of the structural parameters $\bm{\gamma}$, or reward unambiguous beliefs about the world model structure. We try to implement both intrinsic rewards. We denote by $r_{LP}$ the formulation based on learning progress, and by $r_{A}$ the formulation based on ambiguity.

We use Monte Carlo simulations to estimate which action courses would lead to better rewards. The agent samples $N_a$ possible courses of action, and $N_g$ possible world model structures from $p_{\bm{\gamma}}(G)$. For each action courses, the agent simulates the environment dynamics with each sampled model, simulates learning of the structural parameters, and scores the structural parameter update using the chosen reward formulation. For each sampled graph $G$, the structural parameters are updated using as training targets the states predicted through the simulations with graphs $G' \neq G$. As such, courses of actions corresponding to a large disagreement between graphs should lead to larger prediction errors and thus larger learning.

The learning progress reward is then computed as:

\begin{equation}
    r_{LP}(a) = |\Delta_\gamma|
\end{equation}

This intrinsic reward encourages actions towards states that lead to large updates of the agent's current beliefs about the causal structure of its world model. The other implementation we propose is aimed at reducing the ambiguity in the current beliefs, by prioritizing actions leading to states or observations that help reducing the uncertainty about the possible causal relations. It is computed as the average entropy of the Bernoulli distributions on each possible cause-effect relation.

\begin{equation}
    r_{A}(a) = \frac{1}{d_s(d_s+d_a)} \sum_i \sum_k \sigma(\gamma_{ik}') \log \sigma(\gamma_{ik}') + \big( 1 - \sigma(\gamma_{ik}') \big) \log \big( 1 - \sigma(\gamma_{ik}') \big)
\end{equation}

where $\gamma'$ depends on the action course $a$ and corresponds to the updated structural parameters after simulated learning on the trajectories with different graphs $G \sim p_\gamma(G)$.

\section{Experiment}

In order to validate our approach, we need to verify first that the proposed learning methods indeed allow to uncover causal relations in the world model, and second, that the intrinsic motivation functions we have defined accelerate this process.

\begin{figure}[ht!]
    \centering
    \includegraphics[width=0.5\textwidth]{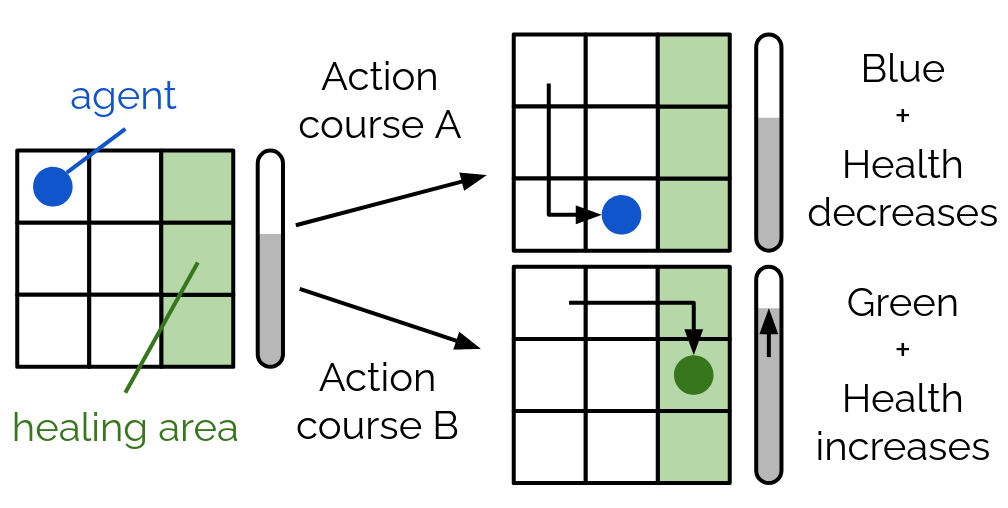}
    \caption{Simulated grid-world environment. Different courses of action can lead to different future states. The agent internally simulates these possible futures and evaluates which action course is more likely to help uncovering the ground-truth causal relations.}
    \label{fig:env}
\end{figure}

We design a very simple experimental setting where the simulated agent evolves in a fully-observable grid-world environment represented in figure \ref{fig:env} (the environment is larger, the size of 3x3 is only used as an illustration). The agent state is composed of its $(X,Y)$ position, a value $X_a$ delimiting a healing area, its health $H$, and its color $C$. Its action space is composed of four possible actions corresponding to the four movement directions $(A_{left}, A_{right}, A_{up}, A_{down})$.

The ground-truth physical mechanisms the agent tries to understand are the following:
\begin{itemize}
    \item The left and right actions affect its $X$ position.
    \item The up and down actions affect its $Y$ position.
    \item If the agent is in the healing area ($X_a\leq X$) then it turns green.
    \item If it is green, then its health increases.
\end{itemize}

In a first stage, we remove the health variable of the agent state, and the agent grasps the first three causal relations using the proposed learning rules, and random actions.

In the second stage, we add the health variable to the agent state, and use the proposed intrinsic motivations to drive the agent. In this stage, the agent has new potential relations where the health variable is the effect, and new potential relations where the health variable is the cause. Initially, the structural parameters for these relations are set to 0, which corresponds after the sigmoid activation to a probability of 0.5 for each relation.

\begin{figure}[ht!]
    \centering
    \includegraphics[width=0.5\textwidth]{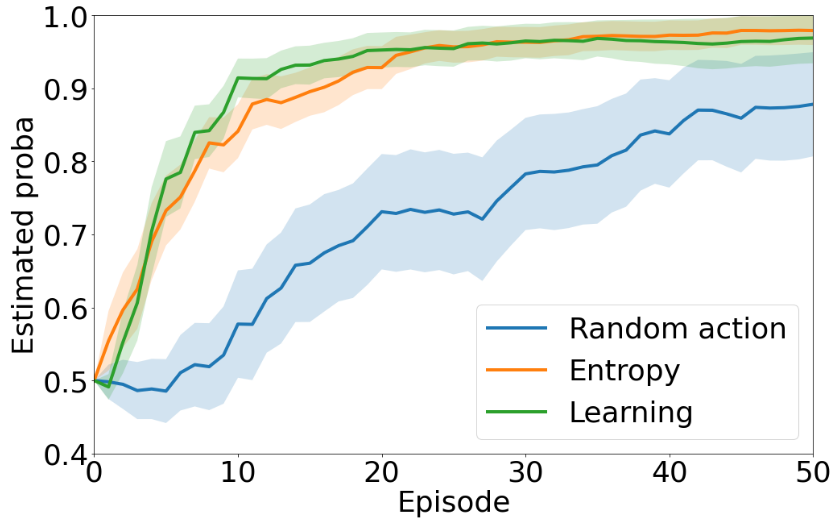}
    \caption{Evolution of the probability that the variable $C$ (color) is a cause for the variable $H$ (health). The lines and areas represent means and standard deviations measured over 100 random seeds.}
    \label{fig:results}
\end{figure}

All these possible causal relations are incorrect, except for the relation setting the health as an effect of the color (the last rule in the previous list). The structural parameters for incorrect relations should decrease, while the structural parameter for the probability that the variable "color" is a cause for the variable "health" should increase. In figure \ref{fig:results} we represent the evolution of this structural parameter using random actions and the two proposed intrinsic motivations.

We can observe that in all three scenarios, the agent belief that color is a cause for health increases. However, the scenario with random actions reach this conclusion in a slower manner. Since the initial agent position can be a few steps to the left of the healing area, action courses where the agent randomly reaches the healing area are not very frequent. Thus, it takes a certain number of episodes for the agent to notice that turning green leads to an increase in health. In the two other scenarios, the agent internally simulates many action courses including some that reach the healing area. These action courses are rated higher, because they allow to test hypothetical causal relations regarding color and health, that are otherwise not tested. In about 10 episodes, the agent has uncovered the correct causal relation between its state variables.

\section{Discussion and future work}

The methods we have used allow to identify different causal mechanisms in the environment, which should provide better generalization to unseen events \citep{bengio2019meta}. Additionally, the intrinsic motivations we have defined improve the sample efficiency of the world model learning algorithm, implementing a wannabe scientist agent that performs experiments on its environment to discriminate between possible theories.

This work is a first step in this direction, and suffers many limitations. The empirical evaluation is simplistic, and the Monte Carlo algorithm we use to evaluate possible action courses could also be improved with a trained policy function. 

It also assumes full observability of the environment with disentangled state features. Interestingly, the inference of a sparse causal graph could also be used as an objective for learning disentangled representations. For instance in our case, if instead we had the agent position defined by the variables $(X+Y, X-Y)$, the agent would need to establish more cause-effect relations to properly predict its future position. Improving the sparsity of the graph should lead to representations that isolate features that can be influenced independently by the agent.

Another limitation comes from the fact that the agent has limited control over the environment. The interventional data processed by the agent only contains interventions on some variables of the environment, allowing to infer some causal relations. However, this does not necessarily allow to discriminate between all the possible causal structures. This is well explained in \citep{wang2022causal}, where the authors suggest categorizing state variables into three groups: variables that can be influenced by the agent, variables that affect the evolution of the variables in the first group but cannot be controlled by the agent, and variables that have no effect on the variables of the first and second groups, and are not controllable by the agent. Since the agent can only intervene on the variables of the first group, other techniques are needed to infer the causal relations of the remaining variables.

\bibliographystyle{unsrtnat}
\bibliography{references}

\end{document}